\documentclass{article}
\usepackage{ijcai24}

\usepackage{times}
\usepackage{soul}
\usepackage{url}
\usepackage[hidelinks]{hyperref}
\usepackage[utf8]{inputenc}
\usepackage[small]{caption}
\usepackage{graphicx}
\usepackage{amsmath}
\usepackage{amsthm}
\usepackage{booktabs}
\usepackage{algorithm}
\usepackage{algorithmic}
\usepackage[switch]{lineno}
\usepackage{amssymb}
\usepackage{color}
\usepackage{epsfig}
\usepackage[accsupp]{axessibility}

\title{Dynamic Brightness Adaptation for Robust Multi-modal Image Fusion}

\author{
Yiming Sun
\and
Bing Cao\thanks{Corresponding author}\and
Pengfei Zhu\And
Qinghua Hu
\affiliations
Tianjin University\\
Tianjin Key Lab of Machine Learning\\
Haihe Lab of ITAI\\
Engineering Research Center of the Ministry of Education on Urban Intelligence and Digital Governance
\emails
\{sunyiming1895,caobing,zhupengfei,huqinghua\}@tju.edu.cn
}

\begin{document}

\maketitle

\begin{abstract}
    Infrared and visible image fusion aim to integrate modality strengths for visually enhanced, informative images.
    Visible imaging in real-world scenarios is susceptible to dynamic environmental brightness fluctuations, leading to texture degradation. Existing fusion methods lack robustness against such brightness perturbations, significantly compromising the visual fidelity of the fused imagery.
    To address this challenge, we propose the Brightness Adaptive multimodal dynamic fusion framework (BA-Fusion), which achieves robust image fusion despite dynamic brightness fluctuations.
    Specifically, we introduce a Brightness Adaptive Gate (BAG) module, which is designed to dynamically select features from brightness-related channels for normalization, while preserving brightness-independent structural information within the source images.
    Furthermore, we propose a brightness consistency loss function to optimize the BAG module. The entire framework is tuned via alternating training strategies.
    Extensive experiments validate that our method surpasses state-of-the-art methods in preserving multi-modal image information and visual fidelity, while exhibiting remarkable robustness across varying brightness levels.
    Our code is available: \url{https://github.com/SunYM2020/BA-Fusion}.
\end{abstract}

\section{Introduction}
Intelligent unmanned systems are susceptible to a decrease in perception ability due to the interference of real dynamic environments~\cite{scirobotics2021abj3947}. Configuring multimodal sensors can effectively enhance their perception ability in complex environments~\cite{Sun2022DroneBasedRC}. Infrared and visible cameras, as a typical set of multimodal sensors, have been widely applied in casualty searching, surveillance missions, $etc.$ However, due to the limitations of hardware devices and imaging mechanisms, visible or infrared cameras can usually only capture partial information of the scene and cannot fully represent the entire scene. Therefore, multimodal image fusion~\cite{Ma2019InfraredAV,Zhang2021ImageFM,Xu2022U2FusionAU} can aggregate the significant contrast information of the infrared modality and the texture detail information of the visible modality, generating fusion images with sufficient information and good visual effects. 
In the past decades, how to design advanced fusion methods has attracted a lot of research attention.

\begin{figure}[!t]
     \centering
     \includegraphics[width=0.48\textwidth]{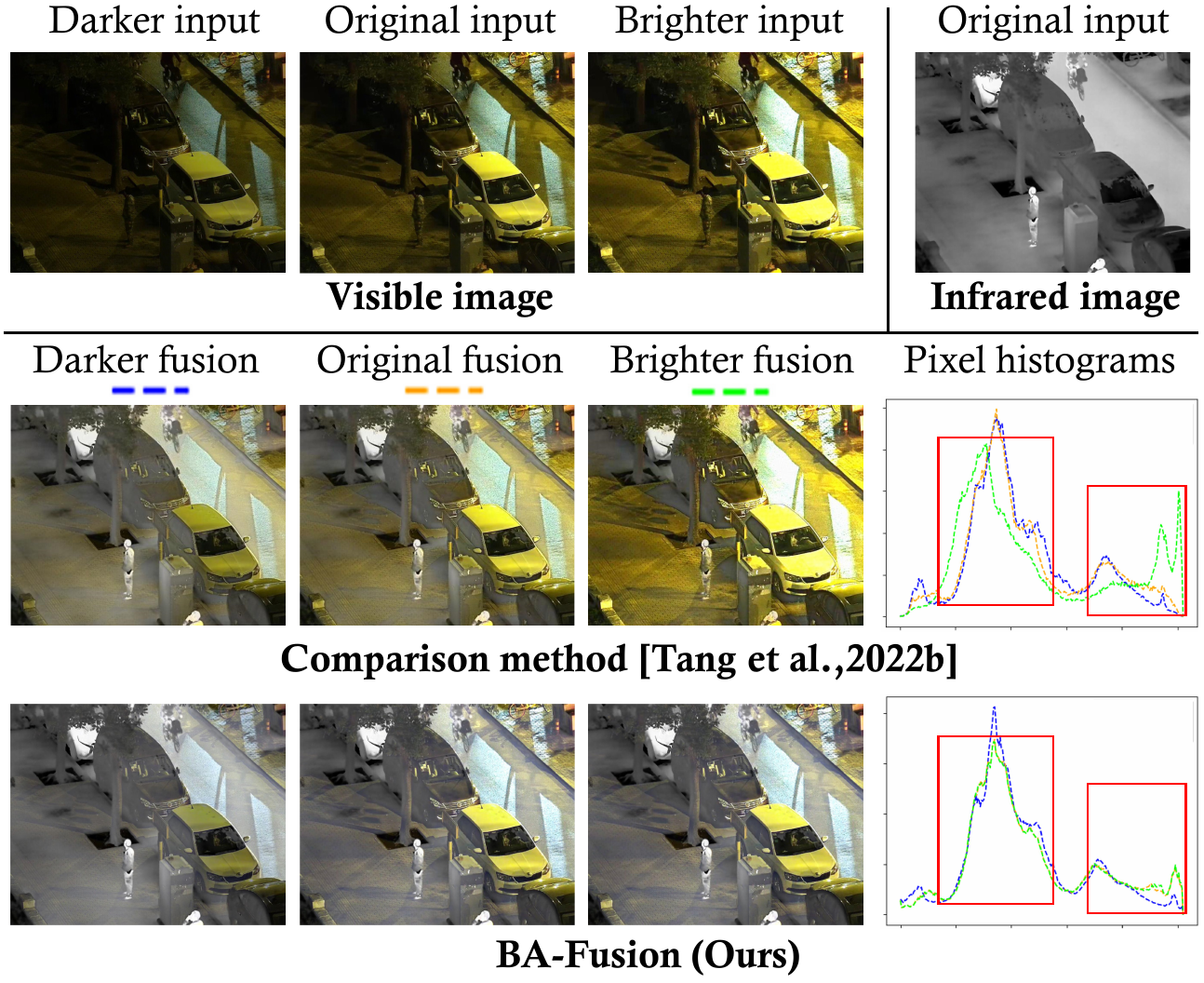}
     \caption{Visual comparisons on fused results and pixel histograms under dynamic brightness conditions. Our method keeps robust performance under varying levels of brightness. }
    \label{fig:intro1_light}
\end{figure}

Existing infrared and visible fusion methods can be divided into two categories: traditional methods represented by image decomposition~\cite{Li2020MDLatLRRAN} and sparse representation~\cite{zhang2018sparse}, and methods based on deep learning. Among them, deep learning-based methods can be further categorized into methods based on pre-trained autoencoder~\cite{li2018densefuse,ZhaoDIDFuse2020,li2021rfn}, methods based on GAN~\cite{ma2019fusiongan,ma2020ddcgan}, methods based on CNN~\cite{Tang2022PIAFusion,sun2022detfusion,cao2023multi}, methods based on diffusion models~\cite{zhao2023ddfm}, and methods based on Transformer~\cite{SwinFuse2022,Tang_2022_YDTR}. However, most of these methods directly combined the texture details and object contrasts of different modalities using a fixed correlation pattern, ignoring the dynamic changes in reality. This makes it difficult for the models to achieve dynamically robust adaptive fusion effects when facing fluctuations in environmental brightness. As shown in Fig.~\ref{fig:intro1_light}, the visible image is affected by brightness interference, resulting in overexposed or dark images. Existing methods lack the dynamic adaptive capability to handle brightness changes, inevitably leading to fluctuations in the fusion image quality with varying environmental brightness, thereby reducing the visual fidelity of the fusion image.

In practical applications, infrared and visible image fusion models should possess robust adaptive capabilities to handle variations in environmental brightness to avoid fusion results being compromised due to brightness fluctuations. To fill this gap, we propose an adaptive dynamic fusion framework that can adapt to changes in brightness and achieve robust fusion of multimodal images under dynamic variations in environmental brightness. In Fig.~\ref{fig:intro1_light}, our method shows a more consistent histogram distribution under different brightness conditions, demonstrating the proposed method effectively balances the impact of brightness fluctuations on the model's learning of texture details and contrast information of different modalities, thus achieving the most robust fusion effect.

Specifically, we propose a dynamic image fusion framework with brightness adaptive gating, termed BA-Fusion, which consists of two parts: Brightness Adaptive Gate (BAG) and multimodal fusion backbone network. The multimodal backbone network is composed of an encoder and a decoder, which are used to extract features from the infrared and visible modalities and generate the fusion image, respectively. The BAG module guides the model to dynamically select the most relevant feature channels with respect to brightness variations in a data-driven manner. It performs brightness normalization on these channels to eliminate the impact of brightness, while the brightness-independent channel features continue to be reused to preserve structural detail information. To train the BAG module, we designed a brightness consistency loss function, which serves as a constraint by ensuring the frequency domain brightness representation of fusion results under different brightness perturbations is consistent with that of normal fusion results. The BAG module gradually establishes the connection between brightness variations and feature channels through alternating training strategies. In this way, the proposed BA-Fusion has the capability of brightness-adaptive robust multimodal fusion.

The main contributions of this paper are summarized as follows:
\begin{itemize}
\item We propose a brightness adaptive dynamic image fusion framework, which effectively mitigates the instability issue in fusion effect caused by environmental brightness fluctuations, enabling robust fusion of infrared and visible images under dynamic brightness conditions.
\item We introduce a brightness adaptive gate module that establishes the correspondence between input image brightness and channel feature representation under the constraint of the brightness consistency loss function.
\item The proposed model dynamically balances the advantages of visible and infrared modalities in terms of texture details and contrast. Extensive experiments on multiple infrared-visible datasets clearly demonstrate our superiority from both quantitative and qualitative perspectives.
\end{itemize}

\section{Related Works}
\label{sec:related}
\subsection{Infrared and Visible Image Fusion}
The infrared and visible image fusion task aims to generate fused images containing richer information by learning the multimodal superiority information~\cite{ma2016infrared}. 
DenseFuse~\cite{li2018densefuse} and DIDFuse~\cite{ZhaoDIDFuse2020} use autoencoder to extract multimodal features and combine them according to predefined rules for fusion. 
Recently, some task-driven fusion methods~\cite{sun2022detfusion,liu2022target,TANG202228SeAFusion} establish a bridge between low-level image fusion and high-level visual tasks. Additionally, GAN-based methods~\cite{ma2019fusiongan,ma2020ddcgan}, Transformer-based methods~\cite{SwinFuse2022,Tang_2022_YDTR} and Diffusion models-based method~\cite{zhao2023ddfm} have also gained widespread attention.
Both DIVFusion~\cite{tang2023divfusion} and L2Fusion~\cite{10223183} tackle the low-light fusion problem through the approach of low-light enhancement, without considering the fluctuations in fusion effect caused by dynamic changes in brightness (overexposure or underexposure). PIAFusion~\cite{Tang2022PIAFusion} introduces an illumination-aware sub-network, but its predicted results are only used for the weights of the infrared intensity loss and visible intensity loss during the training phase. Different from coarse-grained image-level illumination-aware networks, our proposed BA-Fusion constructs a channel-level brightness adaptive framework to achieve dynamic normalization of brightness changes at a finer granularity, ensuring stable learning of rich texture details in the visible modality and achieving robust image fusion under dynamic illumination. 

\subsection{Brightness Correction}
Brightness correction helps to improve image contrast and visual appeal. Low-light image enhancement and exposure correction as typical brightness correction tasks have attracted extensive attention from researchers~\cite{lightadapt,Yao_2023_ICCV,afifi2021learning}. 
Low-light image enhancement~\cite{li2018structure,zhang2022structure} aims to improve image visibility and quality under low-light conditions. 
Exposure correction~\cite{huangdeep2022,huang2022exposure1} aims to adjust brightness components (such as illumination and reflectance) from overexposed versions to normal versions. 
STAN~\cite{zhang2022structure} adopts a divide-and-conquer strategy to model the structural representation and texture representation of low-light images separately. 
FECNet~\cite{huangdeep2022} explores the information represented by the amplitude and phase of the Fourier Transform in exposure correction and designs an interactive module in both frequency and spatial domains to achieve brightness correction. 
Multimodal image fusion tasks aiming to work round-the-clock are naturally challenged by dynamic changes in illumination.
There is still a gap in how to achieve adaptive robust multimodal fusion based on image inputs with different brightness levels. 
For the first time, we establish the dynamic correspondence between channel-level features and brightness in the multimodal fusion task, achieving brightness-adaptive robust fusion.

\begin{figure}[!t]
     \centering
     \includegraphics[width=0.48\textwidth]{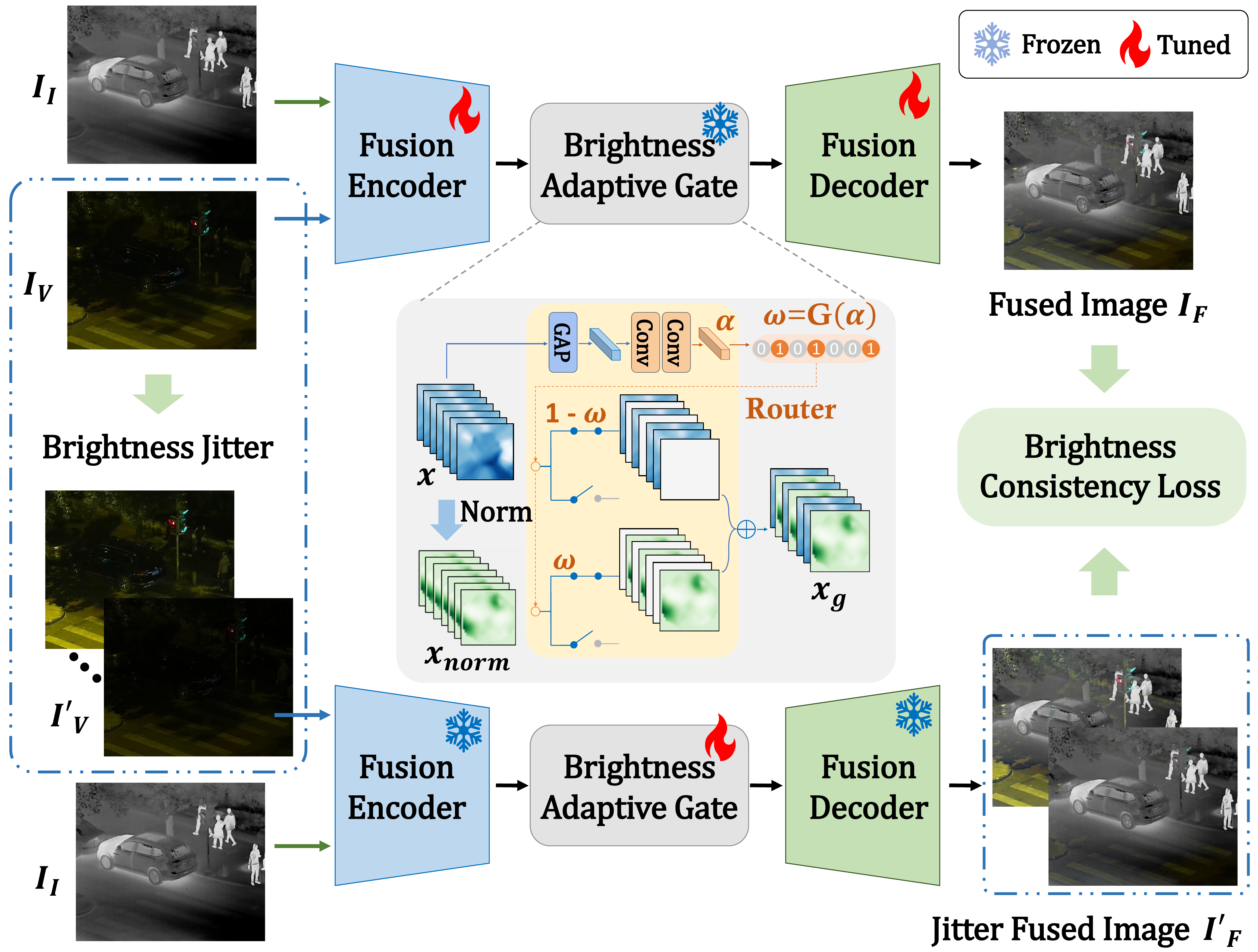}
     \caption{The architecture of BA-Fusion. BA-Fusion consists of a Brightness Adaptive Gate (BAG), and the multimodal fusion backbone network. }
    \label{fig:Core_framework}
\end{figure}

\section{Method}
\subsection{Overall Architecture}
In this paper, we propose a dynamic image fusion framework with brightness adaptive gating, termed BA-Fusion, which contains a Brightness Adaptive Gate (BAG) module and the multimodal fusion backbone network. Fig.~\ref{fig:Core_framework} illustrates the architecture of the proposed BA-Fusion. 
The BAG module dynamically selects original and normalized features along the channel dimensions in a data-driven manner and then combines them to be passed to the decoder.
The multimodal backbone network is composed of an encoder and a decoder, which are used to extract features from the infrared and visible modalities and generate the fusion image, respectively. The structure of the backbone network follows~\cite{chen2022simple}. 
To optimize the BAG module, we also propose an alternating training strategy with a brightness consistency loss function to force the gating module to select the brightness-related channel, which is driven by performance stability under the brightness jitter operation.

In~Fig.~\ref{fig:Core_framework}, a pair of infrared image $I_\mathcal{I} \in \mathbb{R}^{H \times W \times 1}$ and visible image $I_\mathcal{V} \in \mathbb{R}^{H \times W \times 3}$ are fed into the fusion encoders $Enc_\mathcal{F}$ to extract the multi-modal features. Among them, the infrared and visible modalities share the weights of the encoder $Enc_\mathcal{F}$.
The output of the encoder has two parts: the infrared and visible feature maps ($x^\mathcal{I}$ and $x^\mathcal{V}$). Then we feed the visible feature map $x^\mathcal{V}$ to the BAG module to extract the channel-reorganized feature map $x_{g}^\mathcal{V}$. We fuse $x_{g}^\mathcal{V}$ with the infrared feature map $x^\mathcal{I}$ and feed them to the fusion decoder $Dec_\mathcal{F}$, which generates the fused image $I_\mathcal{F} \in \mathbb{R}^{H \times W \times 3}$.  

\subsection{Brightness Adaptive Gate}
As shown in Fig.~\ref{fig:Core_framework}, the BAG module contains two key components, brightness normalization for eliminating brightness-related information and a dynamic gating module for adaptively selecting brightness-related channels. The BAG module selects the feature channels that are most relevant to the brightness change in a data-driven manner guided by dynamic gating. Brightness normalization is then performed on these channels to remove the effects of brightness while continuing to reuse the brightness-independent channel features to preserve structural detail information.

\subsubsection{Brightness Normalization}
Normalization has been proven to eliminate brightness-related components, retain robust structural representations, and effectively reduce the impact of brightness variations on model learning~\cite{huang2022exposure}. We perform channel-wise normalization of the output features $x^\mathcal{V}$ of the encoder $Enc_\mathcal{F}$, the formula of normalization is as follows,
\begin{equation}\label{eq:IN}
    x_{norm}^\mathcal{V}=\gamma \frac{x^\mathcal{V}-\mu(x^\mathcal{V})}{\sigma(x^\mathcal{V})}+\beta,
\end{equation}
where $\mu(x^\mathcal{V})$ and $\sigma(x^\mathcal{V})$ are the mean and standard deviation calculated independently on the spatial dimension of each channel and instance, respectively. $\gamma, \beta\in \mathbb{R}^C$ are learnable parameters. 

The normalized feature $x_{norm}^\mathcal{V}$ has a robust representation irrelevant to brightness changes, which helps the model learn robustly. However, normalization is also a double-edged sword ~\cite{Yao_2023_ICCV}, as it inevitably loses image statistical information and affects image reconstruction accuracy.

\subsubsection{Dynamic Gating Module}
In Fig.~\ref{fig:Core_framework}, we propose a dynamic gate module that takes visible feature maps $x^\mathcal{V}$ as inputs and passes through a set of neural networks to output a routing with a set of binary indicators. This routing is used to guide the model in selecting brightness-related channels, thus achieving dynamic channel selection. Based on the routing selection result, we select only the routed channels from the $x_{norm}^\mathcal{V}$, while retaining the remaining channels of the original feature $x^\mathcal{V}$. Finally, these channel features are recombined to obtain the output feature $x_{g}^\mathcal{V}$. This design effectively mitigates the information loss caused by normalization. The formula of dynamic gate module is as follows,
\begin{equation}\label{eq:gate_interp}
\setlength{\abovedisplayskip}{3pt}
x_{g}^\mathcal{V} = (\mathbf{1}-w)\odot x^\mathcal{V}+w\odot x_{norm}^\mathcal{V},
\setlength{\belowdisplayskip}{3pt}
\end{equation}
where $w$ represents the binary indicators across the channel dimension, and $\odot$ is the channel-wise multiplication. The dynamic gating module contains a global average pooling (GAP) layer with $2$ Conv-ReLu layers, and the output of this module is converted to a binary indicator by the binarization function $G(\alpha)$ to indicate the activation or deactivation of the channel. The $G(\alpha)$ is formalized as,
\begin{equation}\label{eq:gating function}
\setlength{\abovedisplayskip}{1pt}
w=G(\alpha)=\frac{\alpha^2}{\alpha^2+\epsilon},
\setlength{\belowdisplayskip}{1pt}
\end{equation}
where $\epsilon$ is a small positive number. This function transforms $\alpha$ to a value close to one or zero, resulting in an on-off switch gate without requiring additional manual threshold design. 
The introduction of the dynamic gating module enables dynamic normalization of the channels, allowing the recombined features to eliminate the effects of brightness while retaining the basic structural information of the features. 

\subsection{Alternating Training Strategy}\label{sec:joint training}
To train the BAG, we propose an
alternating optimization strategy with brightness consistency loss to drive the dynamic gating module to adaptively select brightness-related channels and retain the brightness-independent channels.

As illustrated in Fig.~\ref{fig:Core_framework}, in the first stage, we train the BA-Fusion model based on the original visible and infrared images. During this stage, the network weights of the BAG module are frozen, and the fusion loss $\mathcal{L}_{fusion}$ is used to optimize the entire multimodal fusion framework. In the second stage, we perform random brightness jitter on the visible images $I_\mathcal{V}$ and input the jittered images $I'_\mathcal{V}$ along with the infrared images $I_\mathcal{I}$ into the BA-Fusion network to generate jittered fusion images $I'_\mathcal{F}$. In this process, to encourage the BAG module to locate and filter out the brightness-related channels, we freeze the model parameters of the fusion encoder $Enc_\mathcal{F}$ and decoder $Dec_\mathcal{F}$ and only optimize the network weights of the BAG module. In the second stage, we propose a brightness consistency loss function, which constrains the consistency of the brightness and structural features of the fusion results under different brightness perturbations with the feature representation of normal fusion results in the first stage. This strategy gradually establishes the connection between brightness variations and feature channels in the BAG module, enabling the fusion model to possess dynamic fusion capability with brightness adaptation. During alternate training, all model weights are shared. 

\subsubsection{Fusion Loss}
According to the alternating training strategy, we divide the network parameters into two groups based on whether they belong to the BAG module and update them using different loss functions. In the first step, we input the original visible and infrared images and update the parameters outside of the BAG module using the fusion loss function. The fusion loss $\mathcal{L}_{fus}$ formula is as follows,
\begin{equation}
\mathcal{L}_{fus}= \mathcal{L}_{pixel}+ \mathcal{L}_{grad}.
\end{equation}

Among them, the pixel loss $\mathcal{L}_{pixel}$ is defined as,
\begin{equation}
\mathcal{L}_{pixel} =\frac{1}{HW} \left \|  I_\mathcal{F} - max\left ( I_\mathcal{V} , I_\mathcal{I} \right ) \right \| _{1},
\end{equation}
where $W$ and $H$ are the width and height. We expect the fused image to preserve the richest texture details of the images from both modalities. So the gradient loss $\mathcal{L}_{grad}$ is formulated as,
\begin{equation}
\mathcal{L}_{grad}=\frac{1}{HW} \left \| \left | \nabla I_\mathcal{F} \right | - max\left ( \left | \nabla I_\mathcal{V} \right |,\left | \nabla I_\mathcal{I} \right | \right ) \right \| _{1},
\end{equation}
where $\nabla$ denotes the Sobel gradient operator, which measures the texture detail information of an image. $\left | \cdot \right |$ stands for the absolute operation.

\subsubsection{Brightness Consistency Loss}
In the second step, we apply brightness jitter to the visible images. During this step, we freeze the parameters outside the BAG module and only update the internal parameters of the BAG using the proposed brightness consistency loss. 
Considering that image brightness is related to the amplitude in the frequency domain~\cite{huangdeep2022}, we introduce frequency domain amplitude information to design the brightness consistency loss. This encourages the network to pay closer attention to the brightness information and effectively select the channels that are related to brightness. The brightness consistency loss $\mathcal{L}_{bcl}$ is formulated as,
\begin{equation}\label{eq:trainigloss2}
\mathcal{L}_{bcl}=|I'_\mathcal{F}-I_\mathcal{F}|_2+|\mathcal{A}(I'_\mathcal{F})-\mathcal{A}(I_\mathcal{F})|_2,
\end{equation}
where $I'_\mathcal{F}$ is the generated jittered fusion images, and $\mathcal{A}$ reprensents the amplitude information in frequency domain. This enables BAG to adaptively select the brightness-related channels to keep the performance on jittered fusion images. 

The two steps are alternately optimized by the $\mathcal{L}_{fus}$ and $\mathcal{L}_{bcl}$, so the overall loss function is formulated as,
\begin{equation}
\mathcal{L}=\mathcal{L}_{fus}+\mathcal{L}_{bcl}.
\end{equation}

\section{Experiments}
\subsection{Experimental Setting}
\paragraph{Datasets and Partition Protocol.} We conducted experiments on two publicly available datasets: (M$^{3}$FD~\cite{liu2022target} and LLVIP~\cite{jia2021llvip}).

\textbf{M$^{3}$FD:} It contains $4,200$ infrared-visible image pairs captured by on-board cameras. We used $3,900$ pairs of images for training and the remaining $300$ pairs for evaluation.

\textbf{LLVIP:} The LLVIP dataset contains $15,488$ aligned infrared-visible image pairs, which is captured by the surveillance cameras in different street scenes. We trained the model with $12,025$ image pairs and evaluated $3,463$ image pairs. 

\paragraph{Competing Methods.}
We compared the $9$ state-of-the-art methods on two publicly available datasets (M$^{3}$FD~\cite{liu2022target} and LLVIP~\cite{jia2021llvip}).  In these comparison methods, DenseFuse~\cite{li2018densefuse} and RFN-Nest~\cite{li2021rfn} are the autoencoder-based methods, PIAFusion~\cite{Tang2022PIAFusion}, DIVFusion~\cite{tang2023divfusion}, and IFCNN~\cite{zhang2020ifcnn} are the CNN-based methods, TarDAL~\cite{liu2022target} is the GAN-based methods. DIDFuse~\cite{ZhaoDIDFuse2020} and DeFusion~\cite{liang2022fusion} are the deep learning-based decomposition methods. YDTR~\cite{Tang_2022_YDTR} is the Transformer-based method.

\paragraph{Implementation Details.}
We performed experiments on a computing platform with four NVIDIA GeForce RTX 3090 GPUs. We used Adam Optimization to update the overall network parameters with the learning rate of $1.0\times 10^{-4}$. The training epoch is set to $60$ and the batch size is $8$.  

\begin{table}
  \renewcommand\tabcolsep{1.8pt}
  \resizebox{\linewidth}{!}{
  \begin{tabular}{cccccccl}
    \toprule
    Method   &SF  &SD  &MI  &VIF  &AG  &$Q_{abf}$\\
    \midrule
        DenseFuse~\cite{li2018densefuse}  &$0.0426$	 &$9.3800$	&$2.6764$	  &$0.6894$	&$3.2640$	 &$0.3093$\\
        IFCNN~\cite{zhang2020ifcnn}     &\textcolor{cyan}{$\mathbf{0.0688}$}	  &\textcolor{cyan}{$\mathbf{9.7633}$}	  &$2.9479$	  &$0.7797$	&\textcolor{cyan}{$\mathbf{5.4136}$}	 &\textcolor{blue}{$\mathbf{0.5845}$}\\
        DIDFuse~\cite{ZhaoDIDFuse2020}    &$0.0550$	  &$7.8074$ 	&$2.5137$	  &$0.5054$	  &$3.4474$	   	&$0.2436$ \\
        RFN-Nest~\cite{li2021rfn}    &$0.0300$	 &$9.7184$	&$2.5042$	  &$0.7294$	 &$2.7853$	  &$0.2287$\\
        PIAFusion~\cite{Tang2022PIAFusion}  &\textcolor{blue}{$\mathbf{0.0787}$}	&$9.7320$	  &\textcolor{cyan}{$\mathbf{3.3690}$}  &\textcolor{blue}{$\mathbf{0.8860}$}	 &\textcolor{red}{$\mathbf{6.0846}$}  &\textcolor{cyan}{$\mathbf{0.5789}$}\\
        YDTR~\cite{Tang_2022_YDTR}     &$0.0474$	  &$8.8701$	  &$2.9152$	   &$0.6322$  &$3.2043$   &$0.2907$\\
        TarDAL~\cite{liu2022target} &$0.0647$  &\textcolor{blue}{$\mathbf{9.7676}$}	&\textcolor{red}{$\mathbf{3.4655}$}	 &$0.7769$  &$4.6094$    &$0.4431$\\
        DeFusion~\cite{liang2022fusion}  &$0.0425$	  &$9.7210$	&$3.2228$	 &$0.7901$    &$3.4194$   &$0.3641$\\
        DIVFusion~\cite{tang2023divfusion}  &$0.0659$	  &$9.7579$	&$2.2851$	 &\textcolor{cyan}{$\mathbf{0.8772}$}    &$5.7428$      &$0.3386$\\
        \bf{BA-Fusion}  &\textcolor{red}{$\mathbf{0.0812}$}	  &\textcolor{red}{$\mathbf{9.8696}$}	&\textcolor{blue}{$\mathbf{3.4454}$}	 &\textcolor{red}{$\mathbf{0.9596}$}    &\textcolor{blue}{$\mathbf{6.0597}$}   &\textcolor{red}{$\mathbf{0.6843}$}\\
  \bottomrule
\end{tabular}}%
  \caption{Quantitative comparison of our BA-Fusion with $9$ state-of-the-art methods on the LLVIP dataset. Bold \textcolor{red}{red} indicates the best, Bold \textcolor{blue}{blue} indicates the second best, and Bold \textcolor{cyan}{cyan} indicates the third best.}
  \label{tab:LLVIPSOTA}
\end{table}

\paragraph{Evaluation Metrics.}
We evaluated the performance of the proposed method based on qualitative and quantitative results. The qualitative evaluation is mainly based on the visual effect of the fused image. A good fused image needs to have complementary information of multi-modal images.
The quantitative evaluation mainly uses quality evaluation metrics to measure the performance of image fusion. We selected $6$ popular metrics, including the spatial frequency (SF)~\cite{Eskicioglu1995ImageQM}, standard deviation (SD), mutual information (MI)~\cite{Qu2002InformationMF}, visual information fidelity (VIF)~\cite{Han2013ANI}, average gradient (AG)~\cite{Cui2015DetailPF}, and gradient-based similarity measurement ($Q_{abf}$)~\cite{Xydeas2000ObjectiveIF}. 
Moreover, we also conducted task-oriented object detection evaluation.

\begin{figure*}[!t]
	\centering
	\includegraphics[width=2.0\columnwidth]{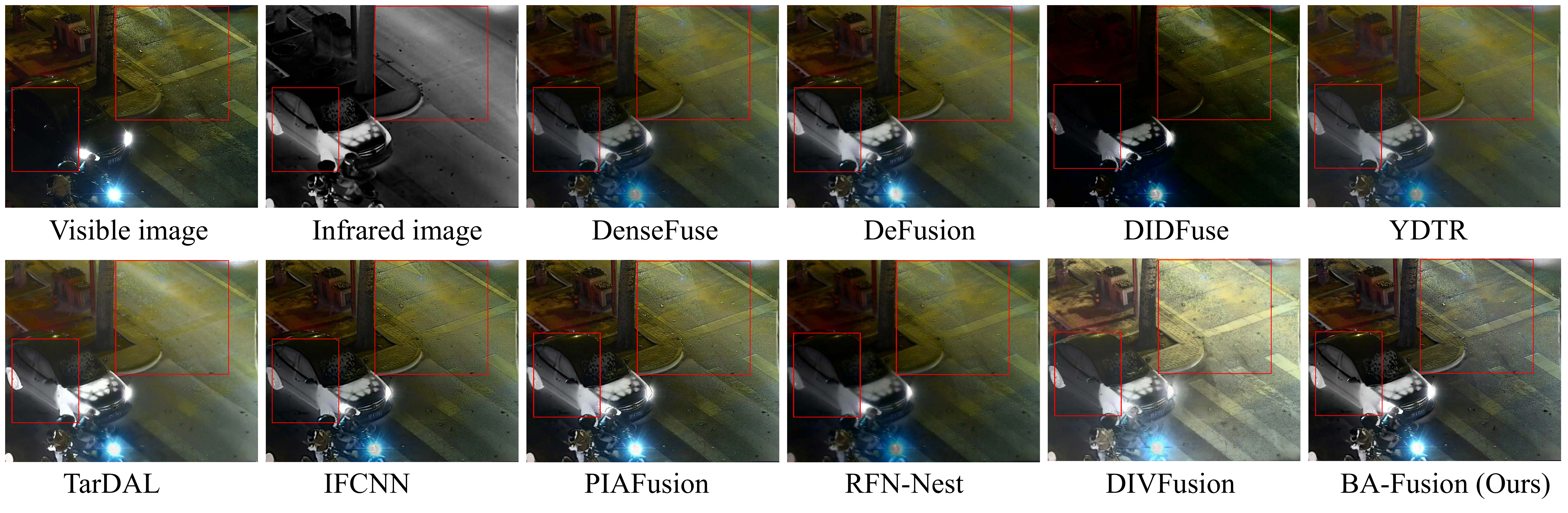}
	\caption{Qualitative comparisons of various methods on representative images selected from the LLVIP dataset.}
	\label{fig:llvipcompare}
\end{figure*}

\subsection{Evaluation on the LLVIP Dataset}
\paragraph{Quantitative Comparisons.}
The quantitative results of the different methods on the LLVIP dataset are reported in~Table~\ref{tab:LLVIPSOTA}. Our method outperforms all the compared methods on $4$ metrics and achieves the second best results on the remaining $2$ metrics, respectively. Specifically, the highest SF we achieved indicates that the proposed method preserves richer texture details in the multi-modal images. As well as the highest SD also indicates that our fusion results can contain the highest contrast information. $Q_{abf}$ denotes the complementary information and edge information transferred from multi-modal images to a fused image, respectively, and our highest results on the metrics indicate that our method can learn more valuable information from multi-modal images. Moreover, the highest VIF also means that our method can generate the most appealing fused images that are more suitable for human vision. These qualitative results demonstrate that BA-Fusion achieves the most superior fusion performance due to its ability to adapt dynamically to brightness.

\paragraph{Qualitative Comparisons.}
According to Fig.~\ref{fig:llvipcompare}, this depicts an actual nighttime road scene with various brightness regions. Our proposed BA-Fusion, with its ability to adapt dynamically to brightness levels, generates the fusion image with the best visual effects. Specifically, for the road traffic markings in the image, our fusion result presents the clearest texture details. On the other hand, the DIVFusion method, which possesses low-light enhancement capability, suffers from severe overexposure in this particular scene, causing blurred ground texture and weak learning of the thermal information of the car body. Furthermore, all competitive methods exhibit some degree of performance degradation in this example. Qualitative comparisons indicate that our method can dynamically fuse images based on their illumination, effectively preserving the advantages of both visible and infrared images, resulting in the fusion image with the highest information content and best visual effects. 

\begin{figure*}[!t]
	\centering
	\includegraphics[width=2.0\columnwidth]{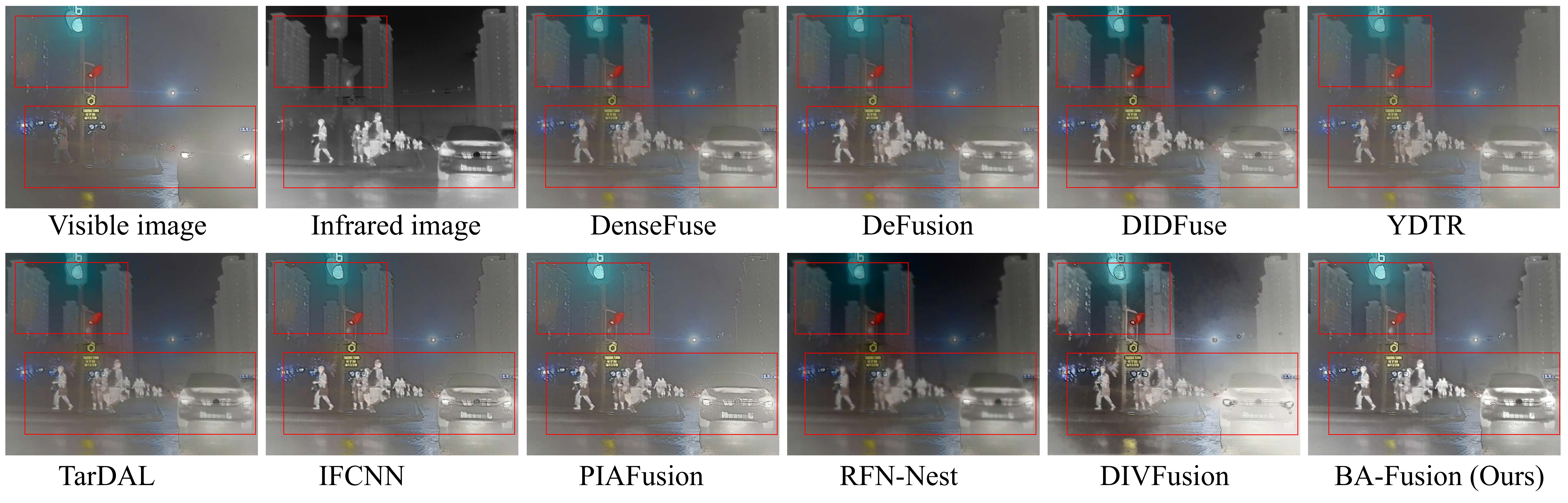}
	\caption{Qualitative comparisons of various methods on representative images selected from the M$^{3}$FD dataset.}
	\label{fig:m3fdcompare}
\end{figure*}

\subsection{Evaluation on the M$^{3}$FD Dataset}
\paragraph{Quantitative Comparisons.}
Table~\ref{tab:M3FDSOTA} presents the results of the quantitative evaluation on the M$^{3}$FD dataset, where our method achieves the best in $4$ metrics and the second best performance in the remaining metrics, respectively. In particular, it shows overwhelming advantages on MI, VIF, and $Q_{abf}$, which indicates that our fusion results contain more valuable information and are more beneficial to the visual perception effect of human eyes. The highest SD and the second-best two metrics (SF \& AG) also show that our fusion results retain sufficient spatial frequency, texture details, and contrast. Benefiting from the dynamic learning of brightness adaptation, our method can learn sufficient texture details and contrast information from both infrared and visible source images, avoiding the interference of visible brightness on the fusion results. Quantitative experiments on M$^{3}$FD also demonstrate the state-of-the-art fusion performance of our proposed framework by dynamic adaptation of brightness.

\paragraph{Qualitative Comparisons.}
As shown in Fig.~\ref{fig:m3fdcompare}, in complex nighttime scenes, visible images suffer from imaging blur and uneven brightness issues, making it difficult for most fusion methods to generate satisfactory results in such scenes. Traditional autoencoder-based methods, such as DenseFuse, DIDFuse, and RFN-Nest, exhibit severe contrast weakening. Additionally, most methods struggle to accurately restore the texture details of cars in the images, as seen in methods like DeFusion, YDTR, and PIAFusion. DIVFusion blindly applies low-light enhancement in this type of scene, leading to severe overexposure issues and fusion failure. Encouragingly, our method achieves the optimal fusion result by leveraging the advantage of adaptive brightness. Our approach effectively preserves the contrast information of pedestrians and cars from the infrared modality while learning all the texture information from the visible images. These qualitative results fully demonstrate that our method dynamically achieves adaptive brightness fusion by establishing the correspondence between brightness and channel features, thereby achieving optimal fusion performance.

\begin{table}
  \renewcommand\tabcolsep{1.8pt}
  \resizebox{\linewidth}{!}{
  \begin{tabular}{cccccccl}
    \toprule
    Method   &SF  &SD  &MI  &VIF  &AG  &$Q_{abf}$\\
    \midrule
        DenseFuse~\cite{li2018densefuse}   &$0.0364$	 &$8.5987$	   &$2.9524$	  &$0.6572$	  &$3.0700$	   &$0.3838$\\
        IFCNN~\cite{zhang2020ifcnn}    &$0.0599$	 &$9.2456$	&$2.9954$	&$0.7522$	&$5.0932$	  &\textcolor{blue}{$\mathbf{0.5755}$} \\
        DIDFuse~\cite{ZhaoDIDFuse2020}     &$0.0420$	  &$9.3409$	  &$2.9955$    &$0.7382$	&$3.5668$	  &$0.4342$ \\
        RFN-Nest~\cite{li2021rfn}      &$0.0345$	&$9.2984$	&$2.9301$	  &$0.7806$	   &$3.1698$	&$0.3772$\\
        PIAFusion~\cite{Tang2022PIAFusion}   &\textcolor{red}{$\mathbf{0.0707}$}	   &\textcolor{blue}{$\mathbf{10.1228}$}   &\textcolor{blue}{$\mathbf{3.8337}$}  &\textcolor{cyan}{$\mathbf{0.8447}$}   &\textcolor{red}{$\mathbf{5.6560}$}    &\textcolor{cyan}{$\mathbf{0.5540}$}\\
        YDTR~\cite{Tang_2022_YDTR}   &$0.0496$	  &$9.2631$	&$3.2128$	 &$0.7276$    &$3.8951$     &$0.4812$\\
        TarDAL~\cite{liu2022target}  &$0.0528$	&$9.6820$	&\textcolor{cyan}{$\mathbf{3.2853}$}	&$0.8347$	&$4.1998$	 &$0.3858$\\
        DeFusion~\cite{liang2022fusion} &$0.0347$	  &$8.9771$	&$2.9505$	 &$0.6752$    &$2.8871$   &$0.3310$\\
        DIVFusion~\cite{tang2023divfusion}  &\textcolor{cyan}{$\mathbf{0.0656}$}	  &\textcolor{cyan}{$\mathbf{9.7279}$}	 &$2.8612$	 &\textcolor{blue}{$\mathbf{0.9569}$}    &\textcolor{cyan}{$\mathbf{5.5956}$}	   &$0.4302$\\
        \bf{BA-Fusion}   &\textcolor{blue}{$\mathbf{0.0687}$}	  &\textcolor{red}{$\mathbf{10.1264}$}	&\textcolor{red}{$\mathbf{3.8431}$}	   &\textcolor{red}{$\mathbf{0.9786}$}    &\textcolor{blue}{$\mathbf{5.6246}$}   &\textcolor{red}{$\mathbf{0.6652}$}\\
  \bottomrule
\end{tabular}}%
  \caption{Quantitative comparison of our BA-Fusion with $9$ state-of-the-art methods on the M$^{3}$FD dataset. Bold \textcolor{red}{red} indicates the best, Bold \textcolor{blue}{blue} indicates the second best, and Bold \textcolor{cyan}{cyan} indicates the third best.}
  \label{tab:M3FDSOTA}
\end{table}

\begin{table}
  \renewcommand\tabcolsep{2pt}
  \resizebox{\linewidth}{!}{
  \begin{tabular}{cccccccl}
    \toprule
    Method   &SF  &SD  &MI  &VIF  &AG  &$Q_{abf}$\\
    \midrule
        w/o BAG    &$0.0810$	 &$9.8555$	 &$3.4241$	 &$0.9553$	  &$6.0504$	  &$0.6825$\\
        w/o BC-Loss     &$0.0811$   &$9.8553$	&$3.4352$	  &$0.9573$    &$6.0585$     &$0.6841$\\
        w/o ATS      &$0.0810$   &$9.8545$	  &$3.4232$	  &$0.9556$	  &$6.0472$	   &$0.6832$\\
        \bf{BA-Fusion}   &$\mathbf{0.0812}$	  &$\mathbf{9.8696}$	&$\mathbf{3.4454}$	 &$\mathbf{0.9596}$    &$\mathbf{6.0597}$   &$\mathbf{0.6843}$\\
  \bottomrule
\end{tabular}}%
  \caption{Ablation studies on LLVIP datasets. BAG denotes \textit{brightness adaptive gating module}, BC-Loss denotes \textit{brightness consistency loss function}, and ATS denotes \textit{alternate training strategy}.}
\label{tab:ablation experiment}
\end{table}

\subsection{Ablation Study}
We construct ablation studies on the LLVIP dataset, which are mainly used to verify the effectiveness of the proposed three core components(brightness adaptive gating module, brightness consistency loss function, and alternate training strategy). The results of the ablation studies are shown in Table~\ref{tab:ablation experiment}.

\paragraph{Brightness Adaptive Gating.}
To verify the effectiveness of Brightness Adaptive Gating (BAG), we removed the brightness adaptive gating module from BA-Fusion and replaced it with a common instance normalization operation, leaving other components unchanged. The experimental results are shown in Table~\ref{tab:ablation experiment}. Without brightness adaptive gating, all evaluation metrics failed to match the performance of BA-Fusion, indicating that normalizing all features may damage texture structure information, leading to a decrease in performance. This also demonstrates that the proposed BAG module is the key to the excellent performance of BA-Fusion.

\paragraph{Brightness Consistency Loss.}
To verify the effectiveness of the Brightness Consistency Loss (BC-Loss), we remove the brightness consistency loss in the alternate training phase, and accordingly, we train the brightness adaptive gating in the second stage using only the fusion loss consistent with the first stage. The experimental results show that removing the brightness consistency loss results in a corresponding overall decrease in the performance of the model, but the magnitude of the decrease is less compared to the other components. This is mainly because brightness constraints can be constructed even if only the fusion loss is used, it is just that these constraints are not as direct and effective as our proposed brightness consistency loss. The experimental results fully demonstrate the effectiveness of the proposed brightness consistency loss.

\paragraph{Alternate Training Strategy.}
To verify the effectiveness of the alternating training strategy, we keep the other components unchanged while canceling the two-step training, we no longer freeze the weights of the different components in BA-Fusion, and we train all components directly. We include the images generated by brightness jitter as part of the dataset and participate in the training of the whole model online, while applying the fusion loss and the brightness consistency loss to the different training image sources. The experimental results show that this mixed online training approach does not achieve the same fusion performance as alternating training, mainly because the fluctuation of image brightness confuses the learning of the model, making it difficult for the model to construct the relationship between brightness and channel features, leading to a degradation of the fusion performance, which also verifies the effectiveness of the proposed alternating training strategy.

\subsection{Visualization}
\begin{figure}[!t]
     \centering
     \includegraphics[width=0.47\textwidth]{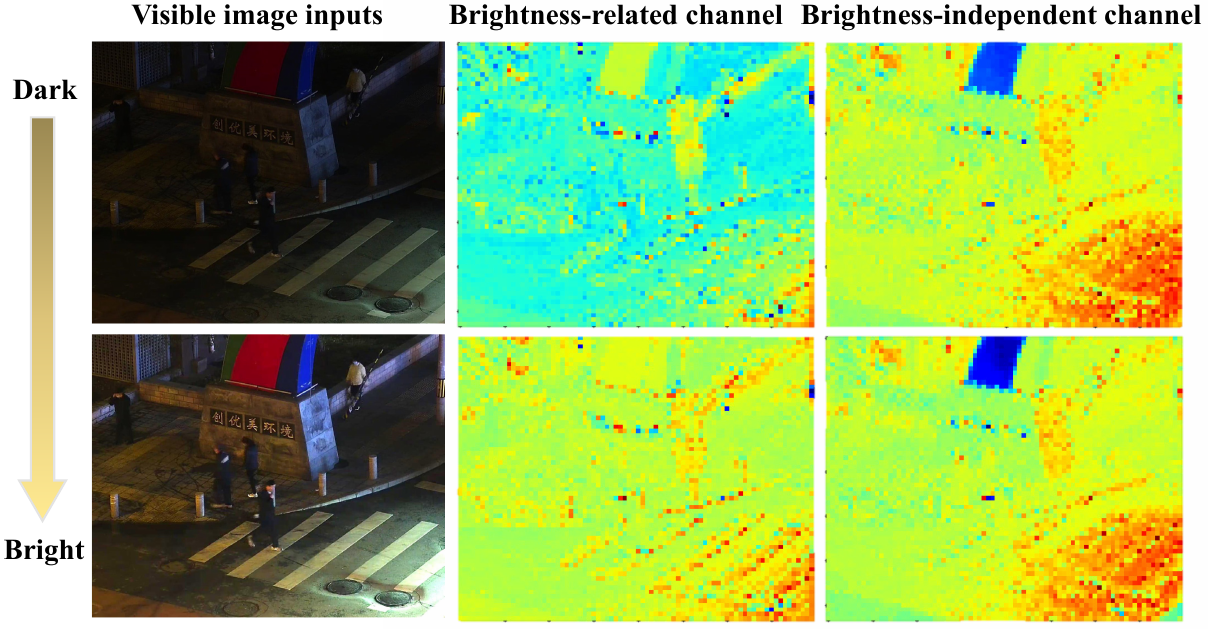}
     \caption{Channel visualization on dynamic brightness conditions. }
    \label{fig: channel}
\end{figure}

We visualize the channel features selected by the BAG module under different brightness conditions. As shown in Fig.~\ref{fig: channel}, the proposed BAG module effectively selects the brightness-related channels and retains the brightness-independent channels. It demonstrates the effectiveness of our method in dynamically perceiving and correcting brightness variations in different regions of the image.

\begin{table}
  \centering
  \resizebox{\linewidth}{!}{
  \begin{tabular}{c|cccccc|c}
    \toprule
    Method   &People  &Car  &Bus  &Motor  &Lamp  &Truck  &mAP\\
    \midrule
        Visible modality  &$68.4$	 &$91.0$	&$91.3$	  &$73.7$	&$81.3$	 &\textcolor{cyan}{$\mathbf{79.0}$} &$80.8$\\
        Infrared modality  &$80.8$	 &$90.1$	&$90.1$	  &$67.7$	&$67.7$	 &$69.0$ &$77.5$\\    
        DenseFuse~\cite{li2018densefuse}  &\textcolor{blue}{$\mathbf{83.9}$}	 &$91.5$	&$90.3$	  &\textcolor{blue}{$\mathbf{75.6}$}	&$82.0$	 &$74.7$   &$83.0$	\\
        IFCNN~\cite{zhang2020ifcnn}     &$83.3$	   &$91.9$	  &$90.5$	  &$74.9$   &\textcolor{cyan}{$\mathbf{86.4}$}	 &$76.9$     &\textcolor{cyan}{$\mathbf{84.0}$}	 \\
        DIDFuse~\cite{ZhaoDIDFuse2020}    &$82.3$	  &$90.9$ 	&$90.1$	  &\textcolor{cyan}{$\mathbf{75.3}$}	  &$83.2$	   	&$76.1$    &$83.0$\\
        RFN-Nest~\cite{li2021rfn}    &$77.7$	 &$90.6$	&\textcolor{cyan}{$\mathbf{91.4}$}	  &$72.0$	 &$79.8$	  &$74.0$    &$80.9$    \\
        PIAFusion~\cite{Tang2022PIAFusion}  &\textcolor{cyan}{$\mathbf{83.5}$}	   &\textcolor{blue}{$\mathbf{92.9}$}	  &$91.3$    &$71.9$	 &$85.4$  &\textcolor{blue}{$\mathbf{80.0}$}  &\textcolor{blue}{$\mathbf{84.2}$}\\
        YDTR~\cite{Tang_2022_YDTR}     &$82.3$	  &\textcolor{cyan}{$\mathbf{92.2}$}	  &$90.9$	   &$73.8$   &\textcolor{blue}{$\mathbf{87.0}$}   &$77.4$   &$83.9$ \\
        TarDAL~\cite{liu2022target}  &$80.9$   &$91.6$	&\textcolor{blue}{$\mathbf{91.5}$}	 &$73.5$   &$81.1$    &$78.7$  &$82.9$	  \\
        DeFusion~\cite{liang2022fusion}   &$83.2$	 &$90.8$	&$89.0$	   &$72.4$    &$76.7$   &$73.7$    &$81.0$   \\
        DIVFusion~\cite{tang2023divfusion}  &$76.4$	  &$90.5$	   &$88.4$	 &$69.1$    &$81.8$      &$78.4$   &$80.8$	 \\
        \bf{BA-Fusion}  &\textcolor{red}{$\mathbf{85.0}$}	  &\textcolor{red}{$\mathbf{93.2}$}	&\textcolor{red}{$\mathbf{91.7}$}	 &\textcolor{red}{$\mathbf{79.3}$}    &\textcolor{red}{$\mathbf{87.1}$}   &\textcolor{red}{$\mathbf{80.4}$} &\textcolor{red}{$\mathbf{86.1}$}\\
  \bottomrule
\end{tabular}}
  \caption{Task-oriented object detection evaluation of our BA-Fusion with $9$ state-of-the-art methods on the M$^{3}$FD dataset. Bold \textcolor{red}{red} indicates the best, Bold \textcolor{blue}{blue} indicates the second best, and Bold \textcolor{cyan}{cyan} indicates the third best. }
  \label{tab:detection}
\end{table}

\subsection{Discussion}
\begin{figure}[!t]
     \centering
     \includegraphics[width=0.48\textwidth]{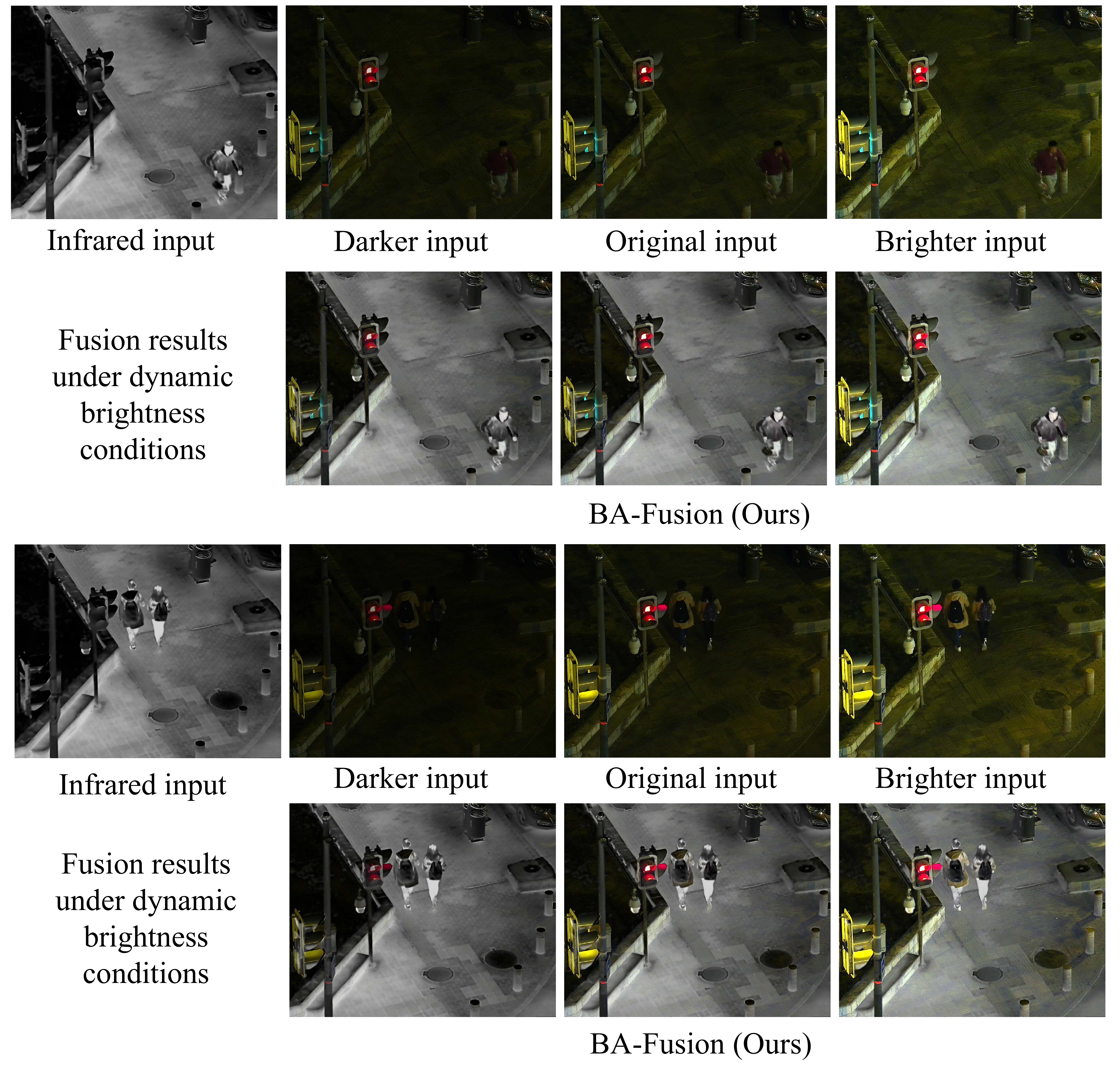}
     \caption{Visual effects on fused result under dynamic brightness conditions. Our brightness adaptive framework (BA-Fusion) keeps the robust performance under varying levels of brightness. }
    \label{fig:testVisual}
\end{figure}

To further verify that our BA-Fusion can realize dynamic robust fusion with brightness adaptation, we use color jitter to generate brighter and darker inputs. As shown in Fig.~\ref{fig:testVisual}, our method can keenly perceive the change of brightness and maintains a consistent fusion effect under varying levels of brightness. 
The proposed method exhibits compatibility with downstream high-level vision tasks, particularly object detection. 
We train the YOLOv5~\cite{glenn_jocher_2020_4154370} model and evaluate the detection performance using mAP as metrics.
The task evaluation results provided in Table~\ref{tab:detection} show that our method achieves the best performance on the object detection task. 

\section{Conclusion}
In this paper, we propose a brightness-adaptive dynamic image fusion framework, BA-Fusion, which consists of a Brightness Adaptive Gate (BAG) module and a multi-modal image fusion backbone network.
The proposed BAG module effectively mitigates the interference of environmental brightness changes on multi-modal image fusion while preserving the structural information of the features. 
Through alternating training strategies, this module establishes a correspondence between image brightness and channel feature representation under the constraints of the proposed brightness consistency loss function, enabling robust multi-modal image fusion under dynamic illumination conditions.
BA-Fusion can dynamically balance the advantages of the two modalities in terms of texture details and contrast, maintaining robust fusion performance in the face of brightness variations.
Experimental results on several challenging datasets demonstrate that the proposed BA-Fusion outperforms state-of-the-art methods in terms of visual effects and quantitative metrics.

\appendix

\section*{Acknowledgments}
This work was supported in part by the National Key R\&D Program of China 2022ZD0116500, in part by the National Natural Science Foundation of China under Grant 62222608, 62106171, and 61925602, in part by the Haihe Lab of ITAI under Grant 22HHXCJC00002, and in part by Tianjin Natural Science Funds for Distinguished Young Scholar under Grant 23JCJQJC00270. This work was also sponsored by CAAI-CANN Open Fund, developed on OpenI Community.

\bibliographystyle{named}
\bibliography{ijcai24}

\end{document}